\documentclass{article}

     \usepackage[final,nonatbib]{neurips_2019}

\usepackage[utf8]{inputenc} 
\usepackage[T1]{fontenc}    
\usepackage{hyperref}       
\usepackage{url}            
\usepackage{booktabs}       
\usepackage{amsfonts}       
\usepackage{nicefrac}       
\usepackage{microtype}      
\usepackage{graphicx}

\title{Taking a Stance on Fake News: Towards Automatic Disinformation Assessment via Deep Bidirectional Transformer Language Models for Stance Detection}

\author{
  Chris Dulhanty, Jason L. Deglint, Ibrahim Ben Daya and Alexander Wong \\
  Systems Design Engineering, University of Waterloo\\
  \texttt{\{chris.dulhanty, jdeglint, ibendaya, a28wong\}@uwaterloo.ca} \\
}

\begin{document}

\maketitle

\begin{abstract}
The exponential rise of social media and digital news in the past decade has had the unfortunate consequence of escalating what the United Nations has called a global topic of concern: the growing prevalence of disinformation\footnote{\url{https://www.ohchr.org/EN/NewsEvents/Pages/DisplayNews.aspx?NewsID=21287&LangID=E}}. Given the complexity and time-consuming nature of combating disinformation through human assessment, one is motivated to explore harnessing AI solutions to automatically assess news articles for the presence of disinformation. A valuable first step towards automatic identification of disinformation is stance detection, where given a claim and a news article, the aim is to predict if the article agrees, disagrees, takes no position, or is unrelated to the claim. Existing approaches in literature have largely relied on hand-engineered features or shallow learned representations (e.g., word embeddings) to encode the claim-article pairs, which can limit the level of representational expressiveness needed to tackle the high complexity of disinformation identification. In this work, we explore the notion of harnessing large-scale deep bidirectional transformer language models for encoding claim-article pairs in an effort to construct state-of-the-art stance detection geared for identifying disinformation. Taking advantage of bidirectional cross-attention between claim-article pairs via pair encoding with self-attention, we construct a large-scale language model for stance detection by performing transfer learning on a RoBERTa deep bidirectional transformer language model, and were able to achieve state-of-the-art performance (weighted accuracy of 90.01\%) on the Fake News Challenge Stage 1 (FNC-I) benchmark. These promising results serve as motivation for harnessing such large-scale language models as powerful building blocks for creating effective AI solutions to combat disinformation.
\end{abstract}

\section{Introduction}

Disinformation presents a serious threat to society, as the proliferation of \textit{fake news} can have a significant impact on an individual's perception of reality. Fake news is a claim or story that is fabricated, with the intention to deceive, often for a secondary motive such as economic or political gain \cite{lazer2018science}. In the age of digital news and social media, fake news can spread rapidly, impacting large amounts of people in a short period of time \cite{vosoughi2018spread}. To mitigate the negative impact of fake news on society, various organizations now employ personnel to verify dubious claims through a manual fact-checking procedure, however, this process is very laborious. With a fast-paced modern news cycle, many journalists and fact-checkers are under increased stress to be more efficient in their daily work. To assist in this process, \textit{automated fact-checking} has been proposed as a potential solution \cite{wu2014toward, ciampaglia2015computational, karadzhov2017fully, hassan2017toward, nadeem2019fakta}.

Automated fact-checking systems aim to assess the veracity of claims through the collection and assessment of news articles and other relevant documents pertaining to the claim at hand. These systems have the potential to augment the work of professional fact-checkers, as well as provide a tool to the public to verify claims they come across online or in their daily lives. An automated fact-checking system consists of several sub-tasks that, when combined, can predict if a claim is truthful \cite{vlachos2014fact}. \textit{Document retrieval} aims to gather relevant articles regarding the claim from a variety of sources. \textit{Stance detection} aims to determine the position of each article with respect to the claim. \textit{Reputation assessment} aims to determine the trustworthiness of each article by analyzing its linguistics and source. \textit{Claim verification} aims to combine stance and reputation information to determine the truthfulness of the claim.

In this paper, we focus on stance detection; given a proposed claim and article, predict if the article agrees, disagrees, has no stance, or is unrelated to the claim. Within the natural language processing (NLP) community, research in stance detection has been catalyzed by the organization of competitions  \cite{mohammad2016semeval, derczynski2017semeval, pomerleau2017fake} and the collection of benchmark datasets \cite{ferreira2016emergent, sobhani2017dataset, chen2019seeing}. Prominent methods addressing stance detection largely differ in terms of their feature representation (e.g., \textit{n}-grams, TF-IDF, word embeddings, etc.) and algorithms (e.g., decision trees, multi-layer perceptions, LSTM networks, etc.); retrospectives on recent challenges \cite{mohammad2016semeval, derczynski2017semeval, hanselowski2018retrospective} provide a comprehensive overview of NLP methods in stance detection. While results have been promising, recent developments in NLP hold the potential for significant improvement. Whereas pre-trained word embeddings such as word2vec \cite{mikolov2013distributed} and GloVe \cite{pennington2014glove} encode language into shallow numerical representations for input to machine learning models, deep bidirectional transformer language models \cite{vaswani2017attention, radford2018improving, devlin2018bert, radford2019language, yang2019xlnet} train on large, unlabelled datasets to learn deeper hierarchical representations of language. The result has been a significant improvement on multi-task NLP benchmarks \cite{wang2018glue}, akin to an "ImageNet moment"\footnote{\url{http://ruder.io/nlp-imagenet/}} for the field.

Motivated by recent advances in NLP and the potential of this technology to meaningfully impact society by addressing the United Nation's Sustainable Development Goals of "Quality Education" and "Peace, Justice, and Strong Institutions"\footnote{\url{https://sustainabledevelopment.un.org}}, we explore the notion of harnessing large-scale deep bidirectional transform language models for achieving state-of-the-art stance detection. Our major contributions are: (1) constructing a large-scale language model for stance detection by performing transfer learning on a RoBERTa deep bidirectional transformer language model by taking advantage of bidirectional cross-attention between claim-article pairs via pair encoding with self-attention, and (2) state-of-the-art results on the Fake News Challenge Stage 1 (FNC-I)\footnote{\url{http://www.fakenewschallenge.org/}} benchmark.

\section{Methodology}

The RoBERTa (Robustly Optimized BERT Approach) model, released in July 2019 by Liu \textit{et al.} \cite{liu2019roberta}, is an open-source language model that achieves state-of-the-art results on benchmark NLP multi-task General Language Understanding Evaluation (GLUE) benchmark \cite{wang2018glue}. RoBERTa is built upon the BERT (Bidirectional Encoder Representations from Transformers) model, released by Devlin \textit{et al.} in October 2018 \cite{devlin2018bert}. RoBERTa and BERT achieve high performance by pretraining a transformer model, initially proposed by Vaswani \textit{et al.} \cite{vaswani2017attention}, in a bidirectional manner on a very large corpus of unlabelled text, and fine-tuning the model on a relatively small amount task-specific labelled data. These models are well-suited for use in stance detection as the pretrained model can be leveraged to perform transfer learning on the target task. Using deep bidirectional transformer language models, RoBERTa and BERT have the ability to gain a deeper understanding of language and context when compared to earlier unidirectional transformer architectures \cite{devlin2018bert}. In addition, RoBERTa demonstrates great results on sentence-pair classification tasks of GLUE, such as Multi-Genre Natural Language Inference \cite{williams2017broad} and Question Natural Language Inference \cite{rajpurkar2016squad, wang2018glue}, tasks very similar in nature to the claim-article classification of stance detection. Following RoBERTa's method of fine-tuning on GLUE tasks, we include both claim and article, separated by a special token, in each example during training and inference.

\section{Experiments and Analysis}

\subsection{Dataset}
To investigate the task of stance detection in the context of fake news detection, we use data released for the Fake News Challenge, Stage 1 (FNC-I). The challenge was organized by Pomerleau and Rao in 2017, with the goal of estimating the stance of an article with respect to a claim. Data is derived from the Emergent dataset \cite{ferreira2016emergent}, sourced from the Emergent Project\footnote{\url{http://www.emergent.info/}}, a real-time rumour tracker created by the Tow Center for Digital Journalism at Columbia University. The stance takes one of four labels: \textbf{Agree} if the article agrees with the claim, \textbf{Disagree} if the article disagrees with the claim, \textbf{Discuss} if the article is related to the claim, but the author takes no position on the subject, and \textbf{Unrelated} if the content of the article is unrelated to the claim. There are approximately 50k claim-article pairs in the training set and 25k pairs in the test set; Table \ref{tab:fc1} summarizes the data distribution.

\begin{table}
  \caption{Statistics of the FNC-I Dataset}
  \label{tab:fc1}
  \centering
  \begin{tabular}{ccc}
    \toprule
         & Training Set     & Test Set \\
    \midrule
    \# of claim-article pairs                   & 49,972    & 25,413\\
    \midrule
    \%  unrelated     & 73.13   & 72.20\\
     \%  discuss     & 17.83   & 17.57\\
     \%  agree     & 7.36   & 7.49\\
     \%  disagree     & 1.68   & 2.74\\
    \bottomrule
  \end{tabular}
\end{table}

\subsection{Metrics}

To evaluate the performance of our method, we report standard accuracy as well as weighted accuracy, suggested by the organizers of the Fake News Challenge, as it provides a more objective metric for comparison given the class imbalance in the dataset. The weighted accuracy, $Acc_w$, is expressed as:

\begin{equation}
Acc_w = 0.25\times Acc_{r, u} + 0.75\times Acc_{a, d, d}
\end{equation}
where $Acc_{r, u}$ is the binary accuracy across related \{agree, disagree, discuss\} and unrelated article-claim pairs, and $Acc_{a, d, d}$ is the accuracy for pairs in related classes only.

\subsection{Model}
\vspace{-0.1in}
We construct our large-scale language model via transfer learning on a pretrained RoBERTa\textsubscript{BASE} deep transformer model, consisting of 12-layers of 768-hidden units, each with 12 attention heads, totalling 125M parameters. We leverage the Transformers library by Hugging Face for implementation \cite{Wolf2019HuggingFacesTS}. To perform transfer learning, we train for three epochs and follow hyperparameter recommendations by Liu \textit{et al.} \cite{liu2019roberta} for fine-tuning on GLUE tasks, namely, a learning rate of 2e-5 and weight decay of 0.1. We train on one NVIDIA 1080Ti GPU with a batch size of 8.

Prior to training, the dataset is pre-processed by initializing each example with a start token to signify the beginning of a sequence, followed by the claim, two separator tokens, the article and an additional separator token. The sequence is then tokenized by RoBERTa's byte-level byte-pair-encoding and trimmed or padded to fit a maximum sequence length of 512. We explore the effects of claim-article pair sequence length and maximum sequence length on classification accuracy in the Appendix.

\subsection{Results \& Discussion}
\vspace{-0.1in}
Results of our proposed method, the top three methods in the original Fake News Challenge, and the best-performing methods since the challenge's conclusion on the FNC-I test set are displayed in Table \ref{tab:res}. A confusion matrix for our method is presented in the Appendix. To the best of our knowledge, our method achieves state-of-the-art results in weighted-accuracy and standard accuracy on the dataset. Notably, since the conclusion of the Fake News Challenge in 2017, the weighted-accuracy error-rate has decreased by 8\%, signifying improved performance of NLP models and innovations in the domain of stance detection, as well as a continued interest in combating the spread of disinformation.

\begin{table}
  \caption{Performance of various methods on the FNC-I benchmark. The first and second groups are methods introduced during and after the challenge period, respectively. Best results are in \textbf{bold}.}
  \label{tab:res}
  \centering
  \begin{tabular}{lll}
    \toprule
     Method     & $Acc_w$ & Acc \\
    \midrule
    Riedel \textit{et al.}  \cite{ucl}                              & 81.72 & 88.46 \\
    Hanselowski \textit{et al.} \cite{athenes}                                            & 81.97 & 89.48\\
    Baird \textit{et al.} \cite{talos}                                 & 82.02 & 89.08\\
    \midrule
    Bhatt \textit{et al.} \cite{Bhatt:2018:CNS:3184558.3191577} & 83.08 & 89.29 \\
    Borges \textit{et al.} \cite{Borges2019}                    & 83.38 & 89.21 \\
    Zhang \textit{et al.} 2018 \cite{Zhang:2018:RMN:3184558.3186919} & 86.66 & 92.00 \\
    Wang \textit{et al.} \cite{Wang:2018:RDD:3184558.3188723}   & 86.72 & 82.91 \\
    Zhang \textit{et al.} 2019 \cite{Zhang:2019:SIH:3308558.3313724} & 88.15 & 93.50 \\
    \midrule
    Proposed Method                                    & \textbf{90.01} & \textbf{93.71} \\
    \bottomrule
  \end{tabular}
\end{table}

\vspace{-0.15in}
\section{Ethical Considerations}
\vspace{-0.15in}
\textbf{Implementation and potential end-users:}\quad The implementation of our stance detection model into a real-world system is predicated on the development of solutions to the document retrieval, reputation assessment and claim verification elements of an automated fact-checking system. While this is an active field of research, it is imperative to note that the reputation assessment sub-task is difficult, as the trustworthiness of an individual or media source may be interpreted differently by different individuals due to personal bias. Provided these elements can be developed, the first intended end-users of an automated fact-checking system should be journalists and fact-checkers. Validation of the system through the lens of experts of the fact-checking process is something that the system's performance on benchmark datasets cannot provide. The implementation of such a system into the daily workflow of these individuals is likely a field of research onto itself. Ultimately, the development of a simple user interface for the general public, such as a browser plug-in, is the goal of this system, assisting individuals to stay informed citizens.

\textbf{Limitations:}\quad The model proposed in this work is limited by the fact that it was trained solely on claims and articles in English, from western-focused media outlets. Further work is necessary to extend this work to other languages, where differences in writing style and cultural norms and nuances may lead to differences in performance. In addition, this model is not designed to deal with satire, where the stance of an article with respect to a claim may appear on the surface to be one way, but the underlying intention of its author is to exploit humor to demonstrate an opposing viewpoint.

\textbf{Risks and potential unintended negative outcomes:}\quad A major risk of a stance detection model or automated fact-checking system is the codification of unintended biases into the model through biased training data. In the field of NLP, gender and racial biases have been reported in word embeddings \cite{bolukbasi2016man, garg2018word} and captioning models \cite{hendricks2018women}; the extent to which such social biases are encoded in recently developed language models is only beginning to be studied \cite{2019arXiv190607337K, NIPS2019_9479}. A secondary risk to the roll-out of these systems for adversarial attacks. Early work by Hsieh \textit{et al.} to investigate the robustness of self-attentive architectures has demonstrated that adversarial examples that could mislead neural language models but not humans are capable of being developed for sentiment analysis, entailment and machine translation \cite{hsieh2019robustness}. In addition, the development of such a system may be interpreted by some as to provide a definitive answer with respect to the truthfulness of a claim, rather than a predictive estimate of its veracity. A potential unintended negative outcome of this work is for people to take the outputs of an automated fact-checking system as the definitive truth, without using their own judgement, or for malicious actors to selectively promote claims that may be misclassified by the model but adhere to their own agenda.

\section{Conclusions}
We have presented a state-of-the-art large-scale language model for stance detection based upon a RoBERTa deep bidirectional transformer. Our promising results motivate efforts to develop additional sub-components of a fully automated fact-checking system such that AI can effectively be harnessed to combat disinformation and allow citizens and democratic institutions to thrive.

\bibliographystyle{unsrt}
\bibliography{main.bib}

\newpage
\appendix

\section{Claim-Article Pair Sequence Length}

Table \ref{tab:tok_len} presents the results of the RoBERTa model on the FNC-I test set, based on the length of claim-article pair. The  model has a maximum sequence length of 512 tokens, so any examples longer than this are trimmed. We find that the model performs best for examples that utilize the full capacity of the input sequence (385 to 512 tokens). Very short sequences (<129 tokens) provide the least amount of information to the model, and the model performs poorly. Long sequences (>512 tokens) have some of their context removed from their input, and these examples also perform relatively poor.

\begin{table}[h]
  \caption{Effect of claim-article pair sequence length of FNC-I test set on classification accuracy of RoBERTa model, with a maximum sequence length of 512.}
  \label{tab:tok_len}
  \centering
  \begin{tabular}{ccc}
    \toprule
    Number of Tokens in Example & Acc & Number of Examples\\
    \midrule
        <129    & 92.05  & 2904  \\
     129-256    & 93.90  & 3606  \\
     257-384    & 95.07  & 6328  \\
     385-512    & \textbf{95.11}  & 4763  \\
        >512    & 92.23  & 7812  \\
    \midrule
    All         & 93.71 & 25413 \\
    \bottomrule
  \end{tabular}
\end{table}

\section{Maximum Sequence Length}

Table \ref{tab:mod_len} presents the results of RoBERTa models of varying maximum sequence lengths on the FNC-I test set. We find an increase in accuracy with a longer maximum sequence length, as more context is provided to the model. We cannot increase the length of the input sequence beyond 512 tokens without training the RoBERTa model from scratch, which is not feasible for us.

\begin{table}[h]
  \caption{Effect of maximum sequence length of RoBERTa model on weighted accuracy and classification accuracy.}
  \label{tab:mod_len}
  \centering
  \begin{tabular}{ccc}
    \toprule
     Maximum Number of Tokens     & $Acc_w$ & Acc \\
    \midrule
    128 & 89.52 & 93.46 \\
    256 & 89.54 & 93.48 \\
    512 & \textbf{90.01} & \textbf{93.71} \\
    \bottomrule
  \end{tabular}
\end{table}

\section{Confusion Matrices}

\begin{figure}[h]
    \centering
    \begin{minipage}{0.49\textwidth}
    \includegraphics[width=0.95\textwidth]{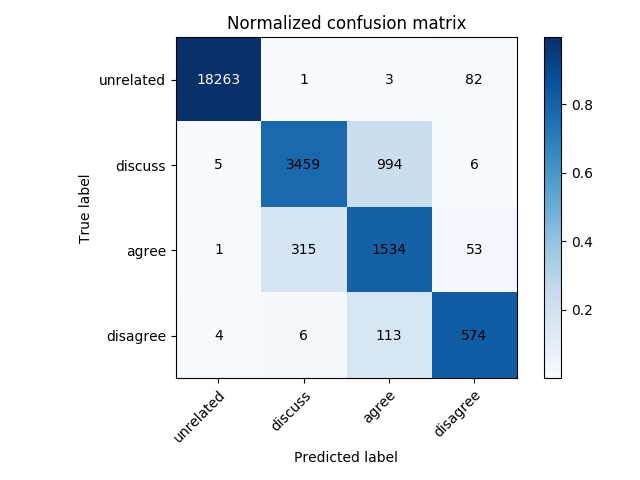}
        \caption{Confusion matrix for Zhang \textit{et al.} \cite{Zhang:2019:SIH:3308558.3313724}.}
        \label{tab:cm_theirs}

    \end{minipage}\hfill
    \begin{minipage}{0.49\textwidth}
    \centering
        \includegraphics[width=0.95\textwidth]{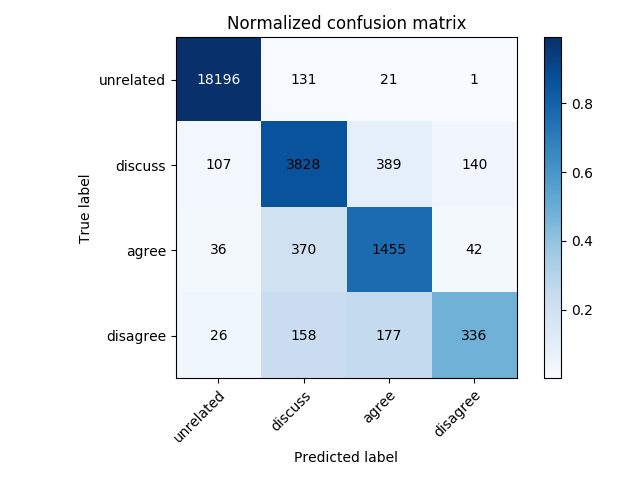}
        \caption{Confusion matrix for proposed method.}
        \label{tab:cm_ours}
    \end{minipage}
\end{figure}

Figures \ref{tab:cm_theirs} and \ref{tab:cm_ours} present confusion matrices for the previous best method and our proposed method on the FNC-I test set.

\end{document}